\documentclass[letterpaper, 10 pt, conference]{ieeeconf}
\IEEEoverridecommandlockouts
\usepackage{graphicx}
\usepackage{amsmath}
\usepackage{amssymb}
\usepackage{booktabs}
\usepackage{multirow}
\usepackage{array}
\usepackage{tikz}
\usetikzlibrary{shapes.geometric, arrows.meta, positioning}
\usepackage{url}
\usepackage{xcolor}
\usepackage{hyperref}
\usepackage{orcidlink}
\hypersetup{colorlinks=true, citecolor=black, linkcolor=black, urlcolor=blue}

\title{\LARGE \bf
Demographic-Aware Transfer Learning for Sleep Stage 
Classification in Clinical Polysomnography
}

\author{S~M~Asif~Hossain\,\orcidlink{0009-0002-4749-109X} and Shruti~Kshirsagar\,\orcidlink{0000-0002-7191-4251}%
\thanks{S~M~Asif~Hossain and Shruti~Kshirsagar are with the School of Computing, Wichita State University, Kansas, USA. E-mails: \href{mailto:sxhossain10@shockers.wichita.edu}{sxhossain10@shockers.wichita.edu} and \href{mailto:shruti.kshirsagar@wichita.edu}{shruti.kshirsagar@wichita.edu}.}}

\begin{document}

\maketitle
\thispagestyle{empty}
\pagestyle{empty}

\begin{abstract}
Automated sleep stage classification typically employs a single population-agnostic model, disregarding established demographic variations in sleep architecture. Sleep patterns, however, differ substantially across gender, age, and obstructive sleep apnea (OSA) severity, indicating that a one-size-fits all approach may be suboptimal for diverse clinical populations. In this paper, we propose a two stage training strategy based on demographic stratification and transfer learning framework. We first  pretrains a convolutional-recurrent model on the full population and then fine tunes it independently for demographic subgroups defined by gender, age, and Apnea-Hypopnea Index (AHI) severity according to the AASM clinical standard. Using the DREAMT dataset comprising 100 clinical subjects and 7 PSG channels, we evaluate 37 fine-tuned configurations across single-axis and two-way demographic combinations. Results demonstrate that 35 of the 37 fine-tuned models outperform the baseline, with Cohen's kappa improvements ranging from 0.9 to 12.9\%. These findings indicate that stratified fine tuning tailored to specific patient demographics yields substantially more accurate sleep staging than a single generalized model, offering a practical and clinically grounded paradigm for personalized sleep assessment.

\end{abstract}

\section{INTRODUCTION}

Sleep is a fundamental physiological process whose disruption
is linked to cognitive decline, cardiovascular morbidity, and
metabolic dysfunction~\cite{irwin2015sleep, chattu2019global};
left unaddressed, such disruption can progress to severe
clinical conditions, underscoring the importance of accurate
and timely sleep disorder diagnosis. Polysomnography (PSG)
remains the clinical gold standard for this purpose, wherein
certified technicians manually classify 30-second epochs into
five stages  Wake, N1, N2, N3, and REM following
AASM guidelines~\cite{berry2012aasm}. Manual scoring,
however, is resource-intensive, subjective, and yields
inter-rater agreement of only approximately
82 - 85\%~\cite{rosenberg2013interscorer, danker2009interrater},
motivating the development of reliable AI-based automated
alternatives.

Deep learning has advanced automated sleep staging
substantially, with convolutional, recurrent, and
attention-based architectures demonstrating strong performance
across benchmark datasets~\cite{supratak2017deepsleepnet,
eldele2021attention, phan2023lseq}. A critical limitation of
existing approaches, however, is their assumption that a single
population-agnostic model generalizes uniformly across
patients  an assumption that is inconsistent with
well-established evidence that sleep architecture varies
systematically across demographic groups. Gender-based
differences in sleep spindle density and EEG spectral power
have been extensively documented~\cite{ohayon2004meta},
age-related changes include progressive loss of slow-wave
activity and increased nocturnal fragmentation~\cite{
ohayon2004meta}, and OSA severity  classified by the AASM
into Normal, Mild, Moderate, and Severe categories based
on the Apnea-Hypopnea Index (AHI)~\cite{berry2012aasm}
 introduces recurrent arousal-driven stage instability that
scales with clinical severity~\cite{gottlieb2020diagnosis,
javaheri2017sleep}. Despite this evidence, no systematic
framework has been proposed to exploit these demographic
differences for improving automated sleep staging within
a single clinical cohort.

This paper aims to close this gap by investigating whether
demographically stratified fine-tuning of a pre-trained sleep
staging model yields consistent and clinically meaningful
improvements over a population-agnostic baseline. 
The key contributions of this work are as follows:

\begin{enumerate}

    \item We propose a two-stage demographic-aware transfer
    learning framework for clinical sleep stage classification,
    wherein a cohort-level pre-trained model is fine-tuned
    independently for subgroups defined by gender, age group,
    and AASM-standard AHI severity, without requiring
    additional recordings or manual feature design.

    \item We provide a systematic evaluation across 37
    single-axis and two-way demographic subgroup
    configurations, demonstrating that 35 of 37 fine-tuned
    models outperform the population-agnostic baseline.

    \item We derive evidence-based deployment guidelines
    identifying which demographic stratification axes yield
    the most consistent gains and establish practical
    subgroup-size thresholds for reliable fine-tuning.

\end{enumerate}

The remainder of this paper is organized as follows: Section~\ref{sec:relatedwork} reviews related work, Section~\ref{sec:method} describes the proposed methodology, Section~\ref{sec:results} presents the results and discussion, and Section~\ref{sec:conclusion} concludes the paper.

\section{RELATED WORKS}
\label{sec:relatedwork}

\subsection{Automated Sleep Staging}

Deep learning has become the dominant approach for automated sleep stage classification. Early convolutional methods demonstrated that discriminative features can be learned directly from raw EEG without manual feature engineering~\cite{sors2018cnn}. Hybrid architectures combining CNNs with recurrent models such as LSTM networks subsequently improved performance by jointly extracting spatial features and modeling temporal transition dynamics~\cite{supratak2017deepsleepnet, mousavi2019sleepeegnet}. Parameter-efficient designs such as TinySleepNet~\cite{supratak2020tinysleepnet} and NanoSleep~\cite{hossain2026nanosleep} showed that competitive accuracy is achievable with substantially fewer parameters. More recently, attention-based and transformer architectures have captured long-range dependencies spanning entire sleep cycles~\cite{eldele2021attention, phan2023lseqsleepnet}, while foundational models trained on large-scale multi-center data have approached expert-level agreement~\cite{fox2025foundational}. Lightweight multi-channel fusion designs have further advanced deployment feasibility~\cite{yang2025lmcsleepnet, fan2024msdcssnet}.

\subsection{Transfer Learning for Sleep Analysis}

Transfer learning has proven effective for adapting across different domains \cite{parupati2025towards,mouradi2026robust, kshirsagar2026geographic, kshirsagar2022cross, tallal2026stda}. Phan et al.~\cite{phan2020transfer} showed that pre-training on large datasets followed by fine-tuning on smaller targets consistently outperforms training from scratch. Van Der Aar et al.~\cite{vanderaar2024transfer} investigated fine-tuning under channel and population mismatches, demonstrating that transfer learning mitigates degradation when models are applied to clinical populations differing from the pre-training cohort. Eldele et al.~\cite{eldele2022transfer} proposed a CNN-RNN transfer framework leveraging pre-trained feature extractors to reduce training time. These studies establish the viability of fine-tuning for cross-dataset adaptation, but focus on bridging domain gaps between different recording environments rather than leveraging demographic structure within a single clinical cohort, which is the focus of the present work.

\subsection{Demographic Variability in Sleep}

The influence of demographic variables on sleep physiology is well established. Dijk~\cite{dijk1989sex} documented gender-based differences in EEG spectral power and spindle characteristics. Ohayon et al.~\cite{ohayon2004meta} conducted a meta-analysis demonstrating age-dependent reductions in slow-wave sleep. Increasing AHI severity has been shown to progressively disrupt sleep macro-architecture~\cite{gottlieb2020osa, tallal2026modulation, sandhuexploring}. Despite this evidence, automated staging systems have not systematically exploited these demographic factors to build specialized models. Our work addresses this gap by proposing a structured framework that explicitly stratifies model training along clinically established demographic axes.

\section{METHODOLOGY}
\label{sec:method}
In this section, we describe the dataset, demographic stratification based on gender, age and AHI severity, model architecture, and evaluation metrics.
\subsection{Dataset}

We utilize the DREAMT v2.1.0 dataset~\cite{wang2024dreamt}, which contains overnight PSG recordings from 100 subjects recruited from the Duke University Health System Sleep Disorders Laboratory. Table~\ref{tab:dataset} summarizes the cohort demographics and sleep stage distribution. This dataset is particularly well-suited for investigating demographic effects on sleep staging because it represents a clinically realistic population rather than a curated healthy cohort. The subjects present a broad spectrum of comorbidities commonly encountered in sleep clinics, including hypertension, diabetes, obesity (68 subjects with BMI $\geq$ 30 kg/m$^2$), and varying severities of obstructive sleep apnea. The age range spans from 21 to 87 years, and AHI values range from near-zero to over 100 events per hour, capturing the full continuum of OSA severity. Certified sleep technicians annotated each 30-second epoch according to AASM guidelines into five categories: Wake (W), N1, N2, N3, and REM. Epochs labeled as Preparation or Missing were excluded. Seven clinically relevant PSG channels were selected: three EEG derivations (C4-M1, F4-M1, O2-M1), two electrooculogram channels (E1, E2), chin electromyography (CHIN), and electrocardiography (ECG). All signals were sampled at 100~Hz, yielding 3000 samples per epoch.

\begin{table}[t]
\caption{Cohort Demographics and Sleep Stage Distribution}
\label{tab:dataset}
\centering
\scriptsize
\setlength{\tabcolsep}{3pt}
\begin{tabular}{@{}l l@{}}
\toprule
\textbf{Characteristic} & \textbf{Value} \\
\midrule
Total subjects & 100 (55 female, 45 male) \\
Age (years) & $56.2 \pm 16.6$ (range: 21--87) \\
BMI (kg/m$^2$) & $33.7 \pm 8.6$ \\
AHI (/h) & $22.1 \pm 28.7$ \\
\midrule
\textbf{AHI severity (AASM)} & \\
\quad Normal (AHI $<$ 5) & 26 subjects \\
\quad Mild (AHI 5--14) & 25 subjects \\
\quad Moderate (AHI 15--29) & 24 subjects \\
\quad Severe (AHI $\geq$ 30) & 25 subjects \\
\midrule
\textbf{Sleep stage epochs} & \\
\quad Wake / N1 / N2 / N3 / REM & 20,041 / 8,818 / 39,953 / 2,704 / 8,387 \\
\quad Total & 79,903 \\
\bottomrule
\end{tabular}
\end{table}

\subsection{Demographic Stratification}

Subjects were stratified along three clinically motivated axes. These axes were selected because they are routinely recorded in clinical PSG and have established associations with sleep architecture.

\textbf{Gender.} The cohort was divided into male ($n{=}45$) and female ($n{=}55$) groups to examine the effect of gender-based differences in sleep EEG morphology on staging accuracy.

\textbf{Age.} Subjects were assigned to one of three age groups: under 50 years ($n{=}33$), 50--65 years ($n{=}34$), and over 65 years ($n{=}33$). The threshold at age~50 corresponds to the onset of menopause-associated hormonal changes that significantly alter sleep architecture ~\cite{ohayon2004meta}, while age~65 aligns with the WHO definition of older adulthood, beyond which age-related EEG slowing and sleep fragmentation become markedly pronounced~\cite{dijk1989sex}.

\textbf{AHI Severity.} Following AASM clinical guidelines~\cite{berry2012aasm, gottlieb2020osa}, subjects were classified into four severity categories based on their overnight AHI: Normal (AHI$<$5, $n{=}26$), Mild (5--14, $n{=}25$), Moderate (15--29, $n{=}24$), and Severe ($\geq$30, $n{=}25$).

In addition to single-axis stratification, two-way combinations were evaluated: Gender$\times$AHI (8 subgroups, $n$ ranging from 8 to 16), Gender$\times$Age (6 subgroups, $n$ from 13 to 21), and Age$\times$AHI (12 subgroups, $n$ from 5 to 12). Three-way stratification was not considered because several resulting subgroups would contain very few subjects.

\subsection{Model Architecture}

\begin{figure*}[t]
    \centering
    \includegraphics[width=\textwidth]{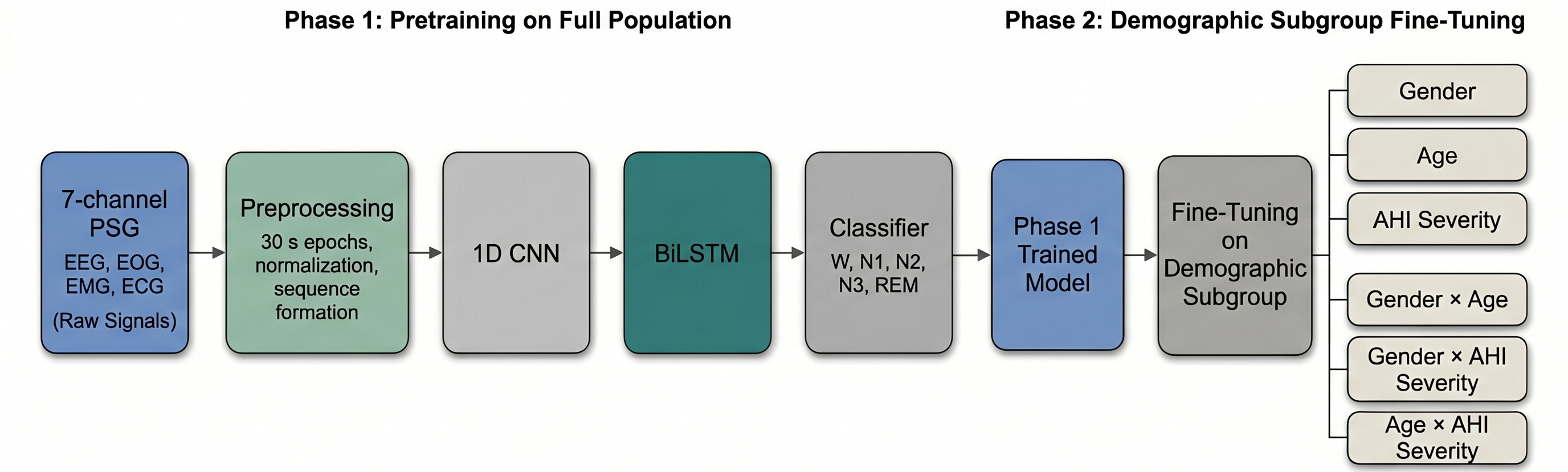}
    \caption{Architecture of the proposed sleep staging model. Phase~1 pre-trains on all 100 subjects. Phase~2 loads Phase~1 weights and fine-tunes on a demographic subgroup with a reduced learning rate.}
    \label{fig:architecture}
\end{figure*}

The classification model, illustrated in Fig.~\ref{fig:architecture}, comprises two components. The feature extractor consists of four one-dimensional convolutional blocks, each containing a convolution layer, batch normalization, ReLU activation, and max-pooling, with progressively increasing filter counts of 64, 128, 128, and 256 \cite{avila2019speech}. A dropout rate of 0.5 is applied after the final convolutional block. Each 30-second epoch is thereby transformed into a compact 256-dimensional feature vector. These feature vectors, computed for each epoch in a sequence of 20 consecutive epochs, are subsequently processed by a two-layer bidirectional long short-term memory (BiLSTM) network with 128 hidden units per direction \cite{kshirsagar2022quality}. The BiLSTM captures temporal dependencies and sleep stage transition dynamics across the epoch sequence.The implementation and experiment code are publicly available at: https://github.com/smAsifHossain/Demographic-Stratified-Sleep-Stage-Classification-via-Transfer-Learning.

\subsection{Architecture Selection}

To select the optimal recurrent backbone, three candidate architectures were evaluated on the full 100-subject cohort under identical training conditions, as summarized in Table~\ref{tab:arch_selection}. The CNN+BiLSTM configuration achieves the highest scores on all three aggregate metrics and exhibits the lowest cross-fold variance ($\kappa$ std of 0.015), indicating the most stable generalization. Bidirectional temporal context is particularly beneficial for sleep staging because both preceding and following epochs carry discriminative information about stage transitions. Based on these results, CNN+BiLSTM was selected as the backbone for all subsequent experiments.

\begin{table}[t]
\caption{Architecture Comparison for Baseline Selection}
\label{tab:arch_selection}
\centering
\scriptsize
\setlength{\tabcolsep}{3pt}
\renewcommand{\arraystretch}{1.08}
\begin{tabular}{@{}lcccc@{}}
\toprule
Architecture & Acc & MF1 & $\kappa$ & $\kappa$ std \\
\midrule
CNN+BiLSTM          & \textbf{0.705} & \textbf{0.647} & \textbf{0.589} & \textbf{0.015} \\
CNN+LSTM+BiLSTM     & 0.699 & 0.639 & 0.581 & 0.033 \\
CNN+LSTM             & 0.689 & 0.629 & 0.570 & 0.037 \\
\bottomrule
\end{tabular}
\end{table}

\subsection{Proposed two-stage training strategy}

Stage 1 (Pre-training):
The model is pre-trained on all 100 subjects using subject-wise 5-fold cross-validation. The 100 subjects are partitioned into five non-overlapping folds of 20 subjects each. In each iteration, four folds (80 subjects) constitute the training set, with 10\% held internally for validation, and the remaining fold (20 subjects) is reserved exclusively for testing. Per-channel z-score normalization statistics are computed from each fold's training set to prevent information leakage. Training employs the Adam optimizer~\cite{kingma2015adam} with an initial learning rate of $10^{-3}$, weighted cross-entropy loss to mitigate class imbalance, gradient clipping at a maximum norm of 5.0, learning rate reduction on plateau (factor 0.5, patience 5 epochs), and early stopping with a patience of 10 epochs. This procedure yields five distinct pre-trained checkpoints, each having never seen its corresponding 20 held-out test subjects during training.

Stage 2 (Demographic fine-tuning)
For each demographic subgroup, the model is fine-tuned using the subgroup's subjects only. Critically, for each fine-tuning fold, the Phase~1 checkpoint is selected such that all test subjects in the current fine-tuning evaluation were also held out during that checkpoint's Phase~1 training. This ensures strict separation between training and evaluation data across both phases, eliminating any possibility of data leakage. The learning rate is reduced to $10^{-4}$ to preserve the broadly learned representations while permitting subgroup-specific adaptation. Single-axis experiments employ 5-fold cross-validation; two-way combinations use 3-fold cross-validation to accommodate smaller subgroup sizes.

\subsection{Evaluation Metrics}
To evaluate performance, all models are assessed using four complementary metrics: 
\begin{enumerate}
  
\item Accuracy (Acc): Accuracy measures the proportion of correctly classified epochs:
\begin{equation}
\mathrm{Acc} = \frac{\sum_{i=1}^{C} TP_i}{N},
\end{equation}
where $C$ is the number of classes, $TP_i$ is the number of true positives for class $i$, and $N$ is the total number of epochs.

\item Macro-F1 Score (MF1):  To account for class imbalance, we report the macro-averaged F1-score, which gives equal importance to each class:
\begin{equation}
F1_i = \frac{2TP_i}{2TP_i + FP_i + FN_i},
\end{equation}
\begin{equation}
\mathrm{MF1} = \frac{1}{C}\sum_{i=1}^{C} F1_i,
\end{equation}
where $FP_i$ and $FN_i$ denote the false positives and false negatives for class $i$.

\item Cohen's Kappa ($\kappa$): Cohen's kappa measures agreement between predicted and reference labels while correcting for chance agreement:
\begin{equation}
\kappa = \frac{p_o - p_e}{1 - p_e},
\end{equation}
where $p_o$ is the observed agreement and $p_e$ is the expected agreement by chance.

\item Per-Class F1-Score: In addition to aggregate metrics, we report the F1-score for each sleep stage (W, N1, N2, N3, and REM) to provide stage-wise performance analysis.

\end{enumerate}

\section{RESULTS AND DISCUSSION}
\label{sec:results}

In this section, we discuss our results and practical implications for clinical deployment.

\subsection{Baseline and Single-Axis Fine-Tuning}

Table~\ref{tab:baseline_single_axis} presents the baseline model alongside all single-axis fine-tuning results. The baseline model, trained on the full cohort of 100 subjects, achieves an accuracy of 0.705, an MF1 of 0.647, and a $\kappa$ of 0.589. All single-axis fine-tuned models outperform this baseline, indicating that subgroup-specific adaptation consistently improves automated sleep staging. This finding is consistent with prior transfer learning studies showing that pre-trained sleep staging models can be effectively adapted to smaller or distribution-shifted cohorts~\cite{phan2020transfer,vanderaar2024transfer,eldele2022transfer}. Among gender-stratified models, the male-specific model achieves $\kappa$ of 0.680, corresponding to a 9.1\% improvement over the baseline. The female-specific model also improves performance, reaching $\kappa$ of 0.629, corresponding to a 4.0\% improvement. These results align with prior physiological evidence that sleep EEG characteristics differ by sex, including spectral power and spindle-related features~\cite{dijk1989sex}. Both gender-specific models also improve REM detection, with REM F1 increasing from 0.733 at baseline to 0.841 for males and 0.792 for females.

\begin{table}[!t]
\caption{Baseline and Single-Axis Fine-Tuning Results Across Gender, Age, and AHI Severity}
\label{tab:baseline_single_axis}
\centering
\scriptsize
\setlength{\tabcolsep}{3pt}
\renewcommand{\arraystretch}{1.08}
\begin{tabular}{lccccccccc}
\toprule
Subgroup & $n$ & Acc & MF1 & $\kappa$ & W & N1 & N2 & N3 & R \\
\midrule
Baseline &   & 0.705 & 0.647 & 0.589 & 0.800 & 0.396 & 0.779 & 0.528 & 0.733 \\
\midrule
\multicolumn{10}{l}{\textit{Gender}} \\
Male     & 45 & 0.772 & 0.683 & 0.680 & 0.852 & 0.496 & 0.820 & 0.403 & 0.841 \\
Female   & 55 & 0.735 & 0.691 & 0.629 & 0.832 & 0.443 & 0.792 & 0.595 & 0.792 \\
\midrule
\multicolumn{10}{l}{\textit{Age}} \\
Over-65    & 33 & 0.753 & 0.619 & 0.652 & 0.867 & 0.477 & 0.795 & 0.209 & 0.747 \\
Under-50   & 33 & 0.742 & 0.700 & 0.637 & 0.795 & 0.441 & 0.802 & 0.675 & 0.789 \\
50--65     & 34 & 0.739 & 0.696 & 0.631 & 0.813 & 0.479 & 0.787 & 0.579 & 0.822 \\
\midrule
\multicolumn{10}{l}{\textit{AHI Severity}} \\
Normal   & 26 & 0.778 & 0.735 & 0.686 & 0.880 & 0.443 & 0.809 & 0.709 & 0.834 \\
Mild     & 25 & 0.754 & 0.684 & 0.649 & 0.800 & 0.462 & 0.806 & 0.507 & 0.847 \\
Moderate & 24 & 0.751 & 0.683 & 0.649 & 0.859 & 0.442 & 0.789 & 0.471 & 0.855 \\
Severe   & 25 & 0.717 & 0.681 & 0.607 & 0.780 & 0.493 & 0.798 & 0.581 & 0.751 \\
\bottomrule
\end{tabular}
\end{table}

Age-stratified fine-tuning also improves performance across all three age groups. The over-65 model achieves the highest age-stratified $\kappa$ of 0.652, corresponding to a 6.3\% improvement, although its N3 F1-score is low at 0.209. This is consistent with established age-related reductions in slow-wave sleep~\cite{ohayon2004meta}. In contrast, the under-50 model achieves an N3 F1-score of 0.675, reflecting the greater preservation of deep sleep in younger adults. Next, AHI-based fine-tuning provides the strongest single-axis improvement. This pattern agrees with clinical evidence that OSA disrupts sleep through repeated respiratory events and arousal-related fragmentation~\cite{gottlieb2020osa,javaheri2017osa}. The Severe group also obtains the highest N1 F1-score among single-axis models, which may reflect the greater prevalence of lighter and fragmented sleep in severe OSA.


\subsection{Gender $\times$ AHI Fine-Tuning}

Table~\ref{tab:genderahi} reports the Gender$\times$AHI fine-tuning results. The best-performing subgroup is M-Mild, which achieves an accuracy of 0.797, an MF1 of 0.756, and $\kappa$ of 0.718. This corresponds to a 12.9\% improvement over the baseline, a 3.8\% improvement over the single-axis Male model, and a 6.9\% improvement over the single-axis Mild model. This suggests that joint stratification can capture interaction effects not fully represented by single-axis models. Most Gender$\times$AHI subgroups improve over the baseline, further supporting demographic-aware transfer learning. This is in line with prior work showing that fine-tuning can reduce performance degradation when target populations differ from the original training cohort~\cite{phan2020transfer,vanderaar2024transfer}. Male subgroups perform better than female subgroups in the Normal, Mild, and Severe AHI categories, while F-Moderate achieves a higher $\kappa$ than M-Moderate. F-Moderate reaches $\kappa$ of 0.652, corresponding to a 6.3\% improvement over the baseline and a 2.3\% improvement over the single-axis Female model. The only Gender$\times$AHI configuration below baseline is F-Severe, with $\kappa$ of 0.521, corresponding to a 6.8\% decrease.

\begin{table}[t]
\caption{Subgroup Fine-Tuning Results Across Gender \texorpdfstring{$\times$}{x} AHI Severity}
\label{tab:genderahi}
\centering
\scriptsize
\setlength{\tabcolsep}{2.5pt}
\renewcommand{\arraystretch}{1.08}
\begin{tabular}{@{}lccccccccc@{}}
\toprule
Subgroup & $n$ & Acc & MF1 & $\kappa$ & W & N1 & N2 & N3 & R \\
\midrule
M-Normal   & 11 & 0.777 & 0.655 & 0.691 & 0.898 & 0.453 & 0.807 & 0.293 & 0.823 \\
M-Mild     & 13 & 0.797 & 0.756 & 0.718 & 0.868 & 0.498 & 0.838 & 0.712 & 0.866 \\
M-Moderate &  8 & 0.739 & 0.602 & 0.618 & 0.795 & 0.466 & 0.819 & 0.128 & 0.803 \\
M-Severe   & 13 & 0.747 & 0.649 & 0.639 & 0.804 & 0.519 & 0.813 & 0.341 & 0.770 \\
\midrule
F-Normal   & 15 & 0.734 & 0.651 & 0.628 & 0.875 & 0.364 & 0.764 & 0.502 & 0.749 \\
F-Mild     & 12 & 0.725 & 0.640 & 0.598 & 0.785 & 0.402 & 0.789 & 0.433 & 0.790 \\
F-Moderate & 16 & 0.750 & 0.710 & 0.652 & 0.878 & 0.465 & 0.772 & 0.608 & 0.828 \\
F-Severe   & 12 & 0.657 & 0.612 & 0.521 & 0.658 & 0.465 & 0.760 & 0.421 & 0.758 \\
\bottomrule
\end{tabular}
\end{table}

\subsection{Gender $\times$ Age Fine-Tuning}

Table~\ref{tab:genderage} presents the Gender$\times$Age results. All three male age subgroups achieve stable improvements, with $\kappa$ values ranging from 0.671 to 0.677, corresponding to improvements of 8.2\% to 8.8\% over the baseline. These values are close to the single-axis Male model, suggesting that additional age stratification provides limited benefit once gender-specific adaptation has already been performed. Female age subgroups show more variability. The female under-50 model improves to $\kappa$ of 0.636, corresponding to a 4.7\% improvement, and achieves the highest N3 F1-score in this group. This is consistent with better preservation of slow-wave sleep in younger adults~\cite{ohayon2004meta}. The female over-65 model also improves over baseline, reaching $\kappa$ of 0.644, corresponding to a 5.5\% improvement. In contrast, the female 50--65 model falls slightly below the baseline with $\kappa$ of 0.580, corresponding to a 0.9\% decrease. The very low N3 F1-score for the male over-65 model also reflects the difficulty of detecting deep sleep in older clinical subjects, where slow-wave sleep is often reduced~\cite{ohayon2004meta}.

\begin{table}[t]
\caption{Subgroup Fine-Tuning Results Across Gender \texorpdfstring{$\times$}{x} Age Groups}
\label{tab:genderage}
\centering
\scriptsize
\setlength{\tabcolsep}{2.5pt}
\renewcommand{\arraystretch}{1.08}
\begin{tabular}{@{}lccccccccc@{}}
\toprule
Subgroup & $n$ & Acc & MF1 & $\kappa$ & W & N1 & N2 & N3 & R \\
\midrule
M-Under-50  & 13 & 0.780 & 0.722 & 0.677 & 0.803 & 0.409 & 0.848 & 0.735 & 0.816 \\
M-50--65    & 13 & 0.767 & 0.676 & 0.671 & 0.844 & 0.467 & 0.815 & 0.419 & 0.836 \\
M-Over-65   & 19 & 0.769 & 0.604 & 0.673 & 0.884 & 0.525 & 0.794 & 0.018 & 0.801 \\
\midrule
F-Under-50  & 20 & 0.739 & 0.700 & 0.636 & 0.789 & 0.438 & 0.807 & 0.751 & 0.714 \\
F-50--65    & 21 & 0.699 & 0.688 & 0.580 & 0.773 & 0.439 & 0.751 & 0.673 & 0.806 \\
F-Over-65   & 14 & 0.751 & 0.621 & 0.644 & 0.861 & 0.410 & 0.818 & 0.249 & 0.767 \\
\bottomrule
\end{tabular}
\end{table}

\subsection{Age $\times$ AHI Fine-Tuning}

Table~\ref{tab:ageahi} shows the Age$\times$AHI results, representing the most granular stratification setting. Ten of the twelve Age$\times$AHI models outperform the baseline despite small subgroup sizes, showing that the pre-trained model can still provide useful initialization for fine-tuning. This is consistent with previous sleep transfer learning studies where pre-training improved performance in smaller target cohorts~\cite{phan2020transfer,eldele2022transfer}.
The strongest Age$\times$AHI result is obtained by the under-50 Normal subgroup, which achieves $\kappa$ of 0.770, accuracy of 0.835, and MF1 of 0.770. This corresponds to an 18.1\% improvement over the baseline, likely reflecting both preserved sleep architecture and the absence of clinically significant apnea. The under-50 Moderate subgroup also achieves the highest N3 F1-score of 0.821, suggesting that younger subgroups may retain distinct stage-specific patterns even when apnea is present. Several older subgroups also benefit from two-way stratification. The over-65 Severe model reaches $\kappa$ of 0.672, corresponding to an 8.3\% improvement, while the over-65 Mild and over-65 Moderate models improve by 3.9\% and 2.3\%, respectively. However, the N3 F1-score of 0.028 for the over-65 Moderate model shows that higher overall agreement does not always imply reliable detection of every stage. This again agrees with prior evidence that slow-wave sleep declines substantially with aging~\cite{ohayon2004meta}. Not all highly granular subgroups benefit. The under-50 Severe and 50--65 Severe models fall below the baseline, with $\kappa$ values of 0.516 and 0.544, corresponding to decreases of 7.3\% and 4.5\%, respectively. These results indicate that severe OSA can reduce fine-tuning reliability when combined with small subgroup size. Clinically, severe OSA is associated with repeated respiratory disturbances and sleep fragmentation~\cite{gottlieb2020osa,javaheri2017osa}, which may weaken stable stage-specific patterns. Thus, two-way stratification can be useful, but should be applied cautiously for very small or severely fragmented subgroups.
\begin{table}[t]
\caption{Subgroup Fine-Tuning Results Across Age \texorpdfstring{$\times$}{x} AHI Severity}
\label{tab:ageahi}
\centering
\scriptsize
\setlength{\tabcolsep}{2.5pt}
\renewcommand{\arraystretch}{1.08}
\begin{tabular}{@{}lccccccccc@{}}
\toprule
Subgroup & $n$ & Acc & MF1 & $\kappa$ & W & N1 & N2 & N3 & R \\
\midrule
Under-50-Normal    &  8 & 0.835 & 0.770 & 0.770 & 0.826 & 0.593 & 0.841 & 0.779 & 0.808 \\
Under-50-Mild      & 11 & 0.729 & 0.658 & 0.616 & 0.720 & 0.366 & 0.851 & 0.501 & 0.716 \\
Under-50-Moderate  &  9 & 0.727 & 0.735 & 0.622 & 0.838 & 0.480 & 0.722 & 0.821 & 0.826 \\
Under-50-Severe    &  5 & 0.657 & 0.677 & 0.516 & 0.788 & 0.367 & 0.822 & 0.598 & 0.563 \\
\midrule
50--65-Normal      &  9 & 0.769 & 0.726 & 0.654 & 0.935 & 0.635 & 0.742 & 0.649 & 0.765 \\
50--65-Mild        &  6 & 0.727 & 0.652 & 0.603 & 0.798 & 0.336 & 0.853 & 0.431 & 0.835 \\
50--65-Moderate    &  5 & 0.753 & 0.688 & 0.652 & 0.885 & 0.421 & 0.721 & 0.419 & 0.797 \\
50--65-Severe      & 14 & 0.662 & 0.652 & 0.544 & 0.723 & 0.452 & 0.747 & 0.520 & 0.853 \\
\midrule
Over-65-Normal     &  9 & 0.728 & 0.707 & 0.622 & 0.878 & 0.430 & 0.794 & 0.700 & 0.774 \\
Over-65-Mild       &  8 & 0.740 & 0.705 & 0.628 & 0.819 & 0.505 & 0.728 & 0.589 & 0.827 \\
Over-65-Moderate   & 10 & 0.728 & 0.582 & 0.612 & 0.856 & 0.425 & 0.865 & 0.028 & 0.927 \\
Over-65-Severe     &  6 & 0.757 & 0.687 & 0.672 & 0.876 & 0.504 & 0.767 & 0.490 & 0.574 \\
\bottomrule
\end{tabular}
\end{table}

\subsection{Practical Implications for Clinical Deployment}
Demographic-aware fine-tuning improves sleep staging without requiring additional sensors or manual feature engineering. Gender, age, and AHI severity are routinely available in clinical sleep studies, enabling deployment using a baseline model and subgroup-specific models selected by patient characteristics~\cite{berry2012aasm,gottlieb2020osa}. Single-axis fine-tuning by gender or AHI severity provides the most consistent improvements, whereas two-way stratification can yield larger gains but is more sensitive to subgroup size and disease severity. These findings support demographic-aware transfer learning for personalized sleep staging. Patient characteristics can guide subgroup-specific model selection in clinical decision-support systems. Small or imbalanced subgroups increase estimation uncertainty, highlighting the need for larger, balanced cohorts to improve deployment reliability and fairness analysis.

\section{CONCLUSION}
\label{sec:conclusion}

This study establishes that a single population-agnostic model is suboptimal for clinical sleep staging across diverse patient demographics. By adopting a two-phase paradigm of pre-training on the full cohort followed by demographic-specific fine-tuning, we achieve consistent improvements across 35 of 37 tested configurations, spanning gender, age, AASM-standard AHI severity, and their pairwise combinations. The proposed approach is clinically grounded in established demographic thresholds, computationally efficient, and directly deployable in clinical settings where patient demographics are routinely available. These findings support a shift toward demographic-aware automated sleep staging, wherein tailored models replace the prevailing one-size-fits all approach, advancing the goal of truly personalized clinical sleep assessment. Future work will evaluate probability calibration and external generalization across independent clinical cohorts and acquisition settings.

\bibliographystyle{ieeetr}
\bibliography{references}

@article{irwin2015sleep,
  author  = {Irwin, Michael R.},
  title   = {Why sleep is important for health: A psychoneuroimmunology perspective},
  journal = {Annual Review of Psychology},
  volume  = {66},
  pages   = {143--172},
  year    = {2015}
}

@article{chattu2019global,
  author  = {Chattu, Vijay Kumar and Manzar, Md. Dilshad and Kumary, Soosanna and Burman, David and Spence, David Warren and Pandi-Perumal, Seithikurippu R.},
  title   = {The global problem of insufficient sleep and its serious public health implications},
  journal = {Healthcare},
  volume  = {7},
  number  = {1},
  pages   = {1},
  year    = {2018}
}

@article{berry2012aasm,
  author  = {Berry, Richard B. and Budhiraja, Rohit and Gottlieb, Daniel J. and Gozal, David and Iber, Conrad and Kapur, Vishesh K. and Marcus, Carole L. and Mehra, Reena and Parthasarathy, Sairam and Quan, Stuart F. and others},
  title   = {Rules for scoring respiratory events in sleep: Update of the 2007 {AASM} manual for the scoring of sleep and associated events},
  journal = {Journal of Clinical Sleep Medicine},
  volume  = {8},
  number  = {5},
  pages   = {597--619},
  year    = {2012}
}

@article{rosenberg2013interscorer,
  author  = {Rosenberg, Richard S. and Van Hout, Steven},
  title   = {The {American Academy of Sleep Medicine} inter-scorer reliability program: Sleep stage scoring},
  journal = {Journal of Clinical Sleep Medicine},
  volume  = {9},
  number  = {1},
  pages   = {81--87},
  year    = {2013}
}

@article{danker2009interrater,
  author  = {Danker-Hopfe, Heidi and Anderer, Peter and Zeitlhofer, Josef and Boeck, Monika and Dorn, Hans and Gruber, Georg and Dorffner, Georg},
  title   = {Interrater reliability for sleep scoring according to the {Rechtschaffen \& Kales} and the new {AASM} standard},
  journal = {Journal of Sleep Research},
  volume  = {18},
  number  = {1},
  pages   = {74--84},
  year    = {2009}
}

@article{sors2018cnn,
  author  = {Sors, Arnaud and Bonnet, St\'{e}phane and Mirek, S\'{e}bastien and Vercueil, Laurent and Payen, Jean-Fran\c{c}ois},
  title   = {A convolutional neural network for sleep stage scoring from raw single-channel {EEG}},
  journal = {Biomedical Signal Processing and Control},
  volume  = {42},
  pages   = {107--114},
  year    = {2018}
}

@article{supratak2017deepsleepnet,
  author  = {Supratak, Akara and Dong, Hao and Wu, Chao and Guo, Yike},
  title   = {{DeepSleepNet}: A model for automatic sleep stage scoring based on raw single-channel {EEG}},
  journal = {IEEE Transactions on Neural Systems and Rehabilitation Engineering},
  volume  = {25},
  number  = {11},
  pages   = {1998--2008},
  year    = {2017}
}

@article{mousavi2019sleepeegnet,
  author  = {Mousavi, Sajad and Afghah, Fatemeh and Acharya, U. Rajendra},
  title   = {{SleepEEGNet}: Automated sleep stage scoring with sequence to sequence deep learning approach},
  journal = {PLoS ONE},
  volume  = {14},
  number  = {5},
  pages   = {e0216456},
  year    = {2019}
}

@inproceedings{supratak2020tinysleepnet,
  author    = {Supratak, Akara and Guo, Yike},
  title     = {{TinySleepNet}: An efficient deep learning model for sleep stage scoring based on raw single-channel {EEG}},
  booktitle = {Proc. 42nd Annual International Conference of the IEEE Engineering in Medicine and Biology Society (EMBC)},
  pages     = {641--644},
  year      = {2020}
}

@article{eldele2021attention,
  author  = {Eldele, Emadeldeen and Chen, Zhenghua and Liu, Chengyu and Wu, Min and Kwoh, Chee-Keong and Li, Xiaoli and Guan, Cuntai},
  title   = {An attention-based deep learning approach for sleep stage classification with single-channel {EEG}},
  journal = {IEEE Transactions on Neural Systems and Rehabilitation Engineering},
  volume  = {29},
  pages   = {809--818},
  year    = {2021}
}

@article{phan2023lseqsleepnet,
  author  = {Phan, Huy and Mikkelsen, Kaare B. and Ch\'{e}n, Oliver Y. and Koch, Philipp and Mertins, Alfred and De Vos, Maarten},
  title   = {{L-SeqSleepNet}: Whole-cycle long sequence modelling for automatic sleep staging},
  journal = {IEEE Journal of Biomedical and Health Informatics},
  volume  = {27},
  number  = {1},
  pages   = {359--370},
  year    = {2023}
}

@article{fox2025foundational,
  author  = {Fox, Benjamin and Jiang, Joy and Wickramaratne, Sajila and Kovatch, Patricia and Suarez-Farinas, Mayte and Shah, Neomi A. and Parekh, Ankit and Nadkarni, Girish N.},
  title   = {A foundational transformer leveraging full night, multichannel sleep study data accurately classifies sleep stages},
  journal = {Sleep},
  volume  = {48},
  number  = {8},
  pages   = {zsaf061},
  year    = {2025}
}

@article{yang2025lmcsleepnet,
  author  = {Yang, Jiayi and Chen, Yuanyuan and Yu, Tingting and Zhang, Ying},
  title   = {{LMCSleepNet}: A lightweight multi-channel sleep staging model based on wavelet transform and multi-scale convolutions},
  journal = {Sensors},
  volume  = {25},
  number  = {19},
  pages   = {6065},
  year    = {2025}
}

@article{fan2024msdcssnet,
  author  = {Fan, Jinghan and Zhao, Mingyang and Huang, Lingyun and Tang, Boren and Wang, Liang and He, Zhenghao and Peng, Xiaogang},
  title   = {Multimodal sleep staging network based on obstructive sleep apnea},
  journal = {Frontiers in Computational Neuroscience},
  volume  = {18},
  pages   = {1505746},
  year    = {2024}
}

@article{phan2020transfer,
  author  = {Phan, Huy and Ch\'{e}n, Oliver Y. and Koch, Philipp and Lu, Zhaojie and McLoughlin, Ian and Mertins, Alfred and De Vos, Maarten},
  title   = {Towards more accurate automatic sleep staging via deep transfer learning},
  journal = {IEEE Transactions on Biomedical Engineering},
  volume  = {68},
  number  = {6},
  pages   = {1787--1798},
  year    = {2021}
}

@article{vanderaar2024transfer,
  author  = {Van Der Aar, Jaap F. and Van Den Ende, Daan A. and Fonseca, Pedro and Van Meulen, Fokke B. and Overeem, Sebastiaan and Van Gilst, Merel M. and Peri, Elisabetta},
  title   = {Deep transfer learning for automated single-lead {EEG} sleep staging with channel and population mismatches},
  journal = {Frontiers in Physiology},
  volume  = {14},
  pages   = {1287342},
  year    = {2024}
}

@article{eldele2022transfer,
  author  = {Eldele, Emadeldeen and others},
  title   = {A deep transfer learning framework for sleep stage classification with single-channel {EEG} signals},
  journal = {Sensors},
  volume  = {22},
  number  = {22},
  pages   = {8826},
  year    = {2022}
}

@article{dijk1989sex,
  title={Sex differences in the sleep EEG of young adults: visual scoring and spectral analysis},
  author={Dijk, Derk Jan and Beersma, Domien GM and Bloem, Gerda M},
  journal={Sleep},
  volume={12},
  number={6},
  pages={500--507},
  year={1989},
  publisher={Oxford University Press}
}

@article{gottlieb2020osa,
  author  = {Gottlieb, Daniel J. and Punjabi, Naresh M.},
  title   = {Diagnosis and management of obstructive sleep apnea: A review},
  journal = {JAMA},
  volume  = {323},
  number  = {14},
  pages   = {1389--1400},
  year    = {2020}
}

@article{javaheri2017osa,
  author  = {Javaheri, Shahrokh and Barbe, Ferran and Campos-Rodriguez, Francisco and Dempsey, Jerome A. and Khayat, Rami and Javaheri, Sogol and Malhotra, Atul and Martinez-Garcia, Miguel A. and Mehra, Reena and Pack, Allan I. and others},
  title   = {Sleep apnea: Types, mechanisms, and clinical cardiovascular consequences},
  journal = {Journal of the American College of Cardiology},
  volume  = {69},
  number  = {7},
  pages   = {841--858},
  year    = {2017}
}

@inproceedings{wang2024dreamt,
  author    = {Wang, Will Ke and Yang, Jiamu and Hershkovich, Leeor and Jeong, Hayoung and Chen, Bill and Singh, Karnika and Roghanizad, Ali R. and Shandhi, Md Mobashir Hasan and Spector, Andrew R. and Dunn, Jessilyn},
  title     = {Addressing wearable sleep tracking inequity: A new dataset and novel methods for a population with sleep disorders},
  booktitle = {Proceedings of the Conference on Health, Inference, and Learning (CHIL)},
  volume    = {248},
  pages     = {380--396},
  year      = {2024}
}

@inproceedings{kingma2015adam,
  author    = {Kingma, Diederik P. and Ba, Jimmy},
  title     = {Adam: A method for stochastic optimization},
  booktitle = {Proceedings of the 3rd International Conference on Learning Representations (ICLR)},
  year      = {2015}
}

@inproceedings{avila2019speech,
  title={Speech-based stress classification based on modulation spectral features and convolutional neural networks},
  author={Avila, Anderson R and Kshirsagar, Shruti R and Tiwari, Abhishek and Lafond, Daniel and O’Shaughnessy, Douglas and Falk, Tiago H},
  booktitle={2019 27th European Signal Processing Conference (EUSIPCO)},
  pages={1--5},
  year={2019},
  organization={IEEE}
}

@article{kshirsagar2022quality,
  title={Quality-aware bag of modulation spectrum features for robust speech emotion recognition},
  author={Kshirsagar, Shruti Rajendra and Falk, Tiago Henrik},
  journal={IEEE Transactions on Affective Computing},
  volume={13},
  number={4},
  pages={1892--1905},
  year={2022},
  publisher={IEEE}
}

@article{kshirsagar2022cross,
  title={Cross-language speech emotion recognition using bag-of-word representations, domain adaptation, and data augmentation},
  author={Kshirsagar, Shruti and Falk, Tiago H},
  journal={Sensors},
  volume={22},
  number={17},
  pages={6445},
  year={2022},
  publisher={MDPI}
}

@inproceedings{parupati2025towards,
  title={Towards Robust Building Damage Detection: Leveraging Augmentation and Domain Adaptation},
  author={Parupati, Bharath Chandra Reddy and Kshirsagar, Shruti and Bagai, Rajiv and Dutta, Atri},
  booktitle={2025 IEEE Green Technologies Conference (GreenTech)},
  pages={163--167},
  year={2025},
  organization={IEEE}
}

@article{mouradi2026robust,
  title={Robust Building Damage Detection in Cross-Disaster Settings Using Domain Adaptation},
  author={Mouradi, Asmae and Kshirsagar, Shruti},
  journal={arXiv preprint arXiv:2603.14694},
  year={2026}
}

@article{kshirsagar2026geographic,
  title={Geographic Bias Analysis and Cross-Domain Generalization in Deep Learning-Based Building Damage Assessment},
  author={Kshirsagar, Shruti and Chandra, Bharath and Tallal, Unaza and Bagai, Rajiv and Dutta, Atri},
  year={2026},
  publisher={Preprints}
}

@article{tallal2026modulation,
  title={Modulation-Based Feature Extraction for Robust Sleep Stage Classification Across Apnea-Based Cohorts},
  author={Tallal, Unaza and Agrawal, Rupesh and Kshirsagar, Shruti},
  journal={Biosensors},
  volume={16},
  number={1},
  pages={56},
  year={2026},
  publisher={MDPI}
}

@article{phan2023lseq,
  author    = {Phan, Huy and Mikkelsen, Kaare B. and 
               Chen, Oliver Y. and Koch, Philipp and 
               Mertins, Alfred and De Vos, Maarten},
  title     = {{L-SeqSleepNet}: Whole-cycle long sequence 
               modelling for automatic sleep staging},
  journal   = {IEEE Journal of Biomedical and Health Informatics},
  volume    = {27},
  number    = {1},
  pages     = {359--370},
  year      = {2023}
}

@article{ohayon2004meta,
  author    = {Ohayon, Maurice M. and Carskadon, Mary A. and 
               Guilleminault, Christian and Vitiello, 
               Michael V.},
  title     = {Meta-analysis of quantitative sleep parameters 
               from childhood to old age in healthy individuals: 
               Developing normative sleep values across the 
               human lifespan},
  journal   = {Sleep},
  volume    = {27},
  number    = {7},
  pages     = {1255--1273},
  year      = {2004}
}

@article{gottlieb2020diagnosis,
  author    = {Gottlieb, Daniel J. and Punjabi, Naresh M.},
  title     = {Diagnosis and management of obstructive sleep 
               apnea: A review},
  journal   = {JAMA},
  volume    = {323},
  number    = {14},
  pages     = {1389--1400},
  year      = {2020}
}

@article{javaheri2017sleep,
  author    = {Javaheri, Shahrokh and Barbe, Ferran and 
               Campos-Rodriguez, Francisco and Dempsey, 
               Jerome A. and Khayat, Rami and Javaheri, 
               Sogol and Malhotra, Atul and 
               Martinez-Garcia, Miguel-Angel and Mehra, 
               Reena and Pack, Allan I. and others},
  title     = {Sleep apnea: Types, mechanisms, and clinical 
               cardiovascular consequences},
  journal   = {Journal of the American College of Cardiology},
  volume    = {69},
  number    = {7},
  pages     = {841--858},
  year      = {2017}
}

@misc{hossain2026nanosleep,
  author       = {Hossain, S. M. Asif and Kshirsagar, Shruti},
  title        = {{NanoSleep}: A Parameter-Efficient Hybrid Temporal Convolutional Network for Single-Channel Sleep Stage Classification},
  year         = {2026},
  month        = jul,
  note         = {Preprint},
  publisher    = {ResearchGate},
  doi          = {10.13140/RG.2.2.23330.39364},
  url          = {https://doi.org/10.13140/RG.2.2.23330.39364}
}

@article{sandhuexploring,
  title={Exploring Explainable AI Methods for Single Channel EEG Sleep Staging Across AHI Stratified Obstructive Sleep Apnea Cohorts},
  author={Sandhu, Gurinder Kaur and Koenig, Jude and Kshirsagar, Shruti and Shukla, Ankita}
}

@article{tallal2026stda,
  title={STDA-Net: Spectrogram-Based Domain Adaptation for cross-dataset Sleep Stage Classification},
  author={Tallal, Unaza and Kshirsagar, Shruti and Shukla, Ankita},
  journal={arXiv preprint arXiv:2605.06736},
  year={2026}
}

\end{document}